\documentclass{article}

\usepackage{amssymb}
\usepackage{graphicx}

\usepackage{stackengine}

\usepackage{makeidx}
\usepackage{graphicx}
\usepackage{amsmath,amssymb} 
\usepackage{color}

\usepackage{epsfig}
\usepackage{graphicx}
\usepackage{amsmath}
\usepackage{amssymb}

\usepackage{mathtools}

\usepackage{relsize}
\usepackage{multirow}
\usepackage{bm}

\usepackage{microtype}
\usepackage{graphicx}
\usepackage{subfigure}
\usepackage{booktabs} 
\usepackage{stackengine}
\usepackage{hyperref}

\usepackage[accepted]{icml2019}

\newcommand{\icol}[1]{
  \begin{smallmatrix}#1\end{smallmatrix}%
}

\icmltitlerunning{Network Parameter Learning with Local Propagation Constraints}

\begin{document}

\twocolumn[
\icmltitle{Network Parameter Learning Using Nonlinear Transforms, Local Representation Goals and Local Propagation Constraints}




\begin{icmlauthorlist}
\icmlauthor{Dimche Kostadinov}{gen} 
\icmlauthor{Behrooz Razeghi}{gen}
\icmlauthor{Slava Voloshynovskiy}{gen}
\end{icmlauthorlist}

\icmlaffiliation{gen}{Department of Computer Science, University of Geneva, Switzerland}

\icmlcorrespondingauthor{Dimche Kostadinov}{dimche.kostadinov@unige.ch}
\icmlcorrespondingauthor{${}$}{dime.kostadinov@gmail.com}


\vskip 0.3in
]



\printAffiliationsAndNotice{}  

\begin{abstract}

In this paper, 
we introduce a novel concept for 
learning of the parameters in a neural network. 
Our idea is grounded on modeling a 
learning problem 
that addresses a trade-off between  (i) satisfying local objectives at each node 
 and (ii) achieving desired data propagation through the network under (iii) local propagation constraints. 
We consider two types of nonlinear transforms which describe the network representations. 
One of the nonlinear transforms serves as activation function. The other one enables a locally adjusted, 
deviation corrective components to be included in the update of the network weights in order to enable attaining target specific representations at the last network node. 
Our learning principle 
not only provides insight into the understanding and the interpretation of the learning dynamics, but it offers theoretical guarantees over 
decoupled and parallel parameter estimation strategy that enables learning in 
synchronous and asynchronous mode. 
Numerical experiments validate the potential of our approach 
on image recognition task. The preliminary results show advantages in comparison to the state-of-the-art methods, w.r.t. the learning time and the network size while having competitive recognition accuracy.
\end{abstract}


\section{Introduction}
\label{sec:introduction}
In the recent years, the multi-layer neural networks (NN) have had significant progress and advances, where impressive results were demonstrated on variety of tasks across many fields \citet{Schmidhuber:2014:Overview}. Addressing estimation/learning of task-relevant, useful and information preserving representation,  the main idea behind the NN learning methods lies in the concept of representing the input through increasingly more abstract layers of feature 
representations.  
Usually, to learn the output of the multi-layer NN representation, a 
target is defined by only one loss (cost) function, which most often is specified in a supervised manner, and is set for the representation at the last node in the network. 
In the most basic case, the problem related to estimation of the parameters in a feed-forward NN 
can be expressed in the following form:
\vspace{-.06in}
\begin{align}
 \! \! \! \!&\{\hat{\bf W}, \hat{\boldsymbol{\theta}}\}=\min_{{\bf W}, \boldsymbol{\theta}}l_f({\bf Y}_0, {\bf W}, \boldsymbol{\theta}) \! =  \min_{{\bf W}, \boldsymbol{\theta}}\sum_{l=1}^{L}l_p({\bf W}_l)+\label{eq:background:problem:formulation} \\
 \! \! \! \!&\sum_{\icol{c=1 \\ k=1}}^{C,K}l_c( f({\bf W}_{L-1} \! \underbrace{f({\bf W}_{L-2}f(...\underbrace{f({\bf W}_1{\bf y}_{0, \{c,k\}})}_{{\bf h}_{1, \{c,k \}}}...))}_{{\bf h}_{L-1, \{c,k \}} }), \boldsymbol{\theta}), \nonumber  
\end{align}
\vspace{-.1in}
{\flushleft where }
$l_f(.,.)$ is a parametric cost function w.r.t. to a certain task objective, \textit{i.e.}, goal, with parameter ${\boldsymbol{\theta} \in \Re^{M_l \times S}}$, ${\bf W}=\{ {\bf W}_1,...,{\bf W}_{L-1} \}$. ${\bf W}_l \in \Re^{M_{l} \times M_{l+1}}$ is the matrix of weights that connects the nodes (layers) at levels $l$ and $l+1$, $f(.)$ is element-wise nonlinear activation function (examples include sigmoid, thanges hyperbolic, exponential, ReLU, etc.), 
and ${\bf h}_{l, \{ c,k \}} \in \Re^{M_l}$ is the network representation at node (layer) level $l$ for the $k$-th input data ${\bf y}_{0, \{c,k\} } \in \Re^{M_1}$ coming from class $c$, and ${\bf Y}_0=[{\bf y}_{0\{1,1\}},{\bf y}_{0, \{1,2\}},...,{\bf y}_{0, \{C,K\} }]$, $k \in \{1,...,K\}, c \in \{1,...,C\}, l \in \{1,...,L \}$.


The cost function 
in \eqref{eq:background:problem:formulation} defines a minimization objective that usually is not convex. In order to estimate the parameters ${\bf W}$ and $\boldsymbol{\theta}$, the most commonly used learning strategy boils down to iterative execution of two steps. In the first step, the data ${\bf y}_{0, \{c,k\} }$ is forward propagated through the network and all ${\bf h}_{l, \{c,k\}}$ are estimated recursively as:
\vspace{-0.07in}
\begin{equation}
\begin{aligned}{\bf h}_{l, \{ c,k\}}=&f({\bf W}_{l-1}{\bf h}_{l-1, \{ c,k \}}), \forall l,c, k.
\label{eq:background:solution:1}
\end{aligned}
\end{equation}
The second step relays on back-propagation \citet{Plaut:1986:BP}, \citet{Lecun::TB:BP} and \citet{Schmidhuber:2014:Overview} with a gradient-based algorithm \citet{LeCun:1998:EBT}, \citet{Yoshua:2012:PR:GBT} in order to update ${\bf W}_l$ and $\boldsymbol{\theta}$ and minimize the non-convex objective in \eqref{eq:background:problem:formulation}. 
At iteration $t$, starting from the last node at level $l=L$ and using the gradient and/or second order derivative information of the objective w.r.t. the weight ${\bf W}_{l}^t$, the parameters ${\bf W}_{l}^t$ are updated sequentially by back propagating the loss through the NN. 
The common update rule has the following form:
\begin{equation}
\begin{aligned}
{\bf W}_{l}^{t+1}={\bf W}_{l}^{t}-&\alpha \frac{\partial l_f( {\bf Y}_0, {\bf W}, \boldsymbol{\theta}) }{\partial {\bf W}_{l}^{t}}+\\
&\beta \mathcal{V}(\frac{\partial l_f( {\bf Y}_0, {\bf W}, \boldsymbol{\theta}) }{\partial {\bf W}_{l}^{t}} , {\bf W}_{l}^{t}, {\bf W}_{l}^{t-1}), 
\label{eq:background:solution:2}
\end{aligned}
\end{equation}
where $\mathcal{V}(\frac{\partial l_f( {\bf Y}_0, {\bf W}, \boldsymbol{\theta}) }{\partial {\bf W}_{l}^{t}} , {\bf W}_{l}^{t}, {\bf W}_{l}^{t-1})$ represent the second order derivative of the objective w.r.t. ${\bf W}_{l}^t$ or it approximation as proposed in \citet{Bottou:2012:SGD_Tricks}, \citet{Shamir:2013:SGD_NON_Smood}, \citet{Srivastava:2014:Dropout} and \citet{Kingma:2014:Adam}.

One of the most crucial issues in the above approach is the second step. Since in the back-propagation a gradient based algorithm is used, the problem of vanishing gradient \citet{Hochreiter:1998:VGP}  or the exploding  gradient \citet{Razvan:Pascanu:Exp:Grad:prob} might lead to a non-desirable local minimum (or saddle point).
On the other hand, the dependencies from the subsequent propagation do not allow parallel parameter learning per node, while an additional challenge is the interpretation of the learning dynamics during training. 

In the last decade, many works \citet{Bottou:2012:SGD_Tricks}, \citet{Shamir:2013:SGD_NON_Smood}, \citet{Srivastava:2014:Dropout},\citet{Kingma:2014:Adam}, \citet{Loshchilov:2016:SGDR}, \citet{Ruder:2016:Overview} \citet{Goh:2017:why_MW},\citet{Zhu:Adam_Look_Ahead} 
have addressed issues related with the gradient based weight update. 
Parallel parameters updates were addressed by 
 \citet{JaderbergCOVGK:2016} and \citet{CzarneckiSJOVK:2017:Sintetic:Gradient} 
and the methods proposed by \citet{LeeZ:2014TP}, \citet{Balduzzi:2015:KCB}, \citet{Taylor:2016BXSPG} and \citet{Nokland:2016DF}. 

While the aforementioned manuscripts provide means to surpass sequential 
updates they still fall within the realm of the concept that uses "propagated information" about the deviations from only one goal (target) at the last network node.  Other alternatives that allow posing local goals on the network representation while enabling parallel update on the network weighs by including a local propagation constraints were not addressed. In this line, a network learning principle that takes into account local correction component that is based on the deviations from a local goals on the representations at a given network node and its closely connected nodes was not explored. 

\subsection{Learning Model Outline} 


In the usual problem  modeling by \eqref{eq:background:problem:formulation}, only one objective function is defined for the representations on the last node in the network and one predefined activation function is used. 

In this paper, in order to allow a local decoupling per the network nodes that enables parallel update of the parameters, as well as provides a possibility to interpret and explain the learning dynamics, 
we describe a novel learning concept. 
In our problem modeling, we introduce (i) two types of nonlinear transforms per network node (ii) a local objective at each node related to the corresponding local representation goal and (iii) a local propagation constraints. 

\subsubsection{Sparsifiying Nonlinear Transform as Activation Function} In the most simple case, analogous to the commonly used description by an activation function \eqref{eq:background:solution:1}, we use a sparsifying nonlinear transforms (sNTs) \citet{RubinsteinE14} and \citet{DBLP:conf/icassp/RavishankarB14}. We denote the sNT representation at node level $l$ defined w.r.t. 
a \textit{sparsifying transform} 
with parameter set $\mathcal{S}_l=\{ {\bf A}_{l-1}, \tau_{l} \}$ ($\tau_{l} \geq 0$ is a thresholding parameter) as:
\vspace{-.0in}
\begin{align}
&{\bf u}_{l,\{ c,k\}}={\rm sign}( {\bf q}_{l, \{c,k\}}) \odot \max( \vert {\bf q}_{l, \{c,k\}} \vert-\tau_{l}{\bf 1}, {\bf 0}), \label{eq:sparse:representation}
\end{align}
where ${\bf q}_{l, \{c,k\}}={\bf A}_{l-1}{\bf u}_{l-1, \{c,k\}}$ is the linear transform, ${\bf u}_{0, \{c,k\}}={\bf y}_{0,
\{ c,k \} }$ and ${\bf A}_{l}\in \Re^{M_{l-1} \times M_{l}}$ is the weight that connects two nodes at levels $l-1$ and $l$.

\subsubsection{Local Representation Goals} By our learning principle, 
we introduce local objectives ({local goals}) per all representations that 
describe the 
desired representations per node, 
which are formally defined w.r.t. a linear transform representation at that node and 
a function. A key here is that we use a function analogous to the concept of objective, but, the difference is that we define the functional mapping as a solution to an optimization problem, where its role is to transform a given representation into a representation with specific properties, \textit{e.g.}, discrimination, information preserving, local propagation constraints preserving, sparsity, compactness, robustness etc. 

\subsubsection{Local Model for Deviation Correction} When we propagate data forward through the network, the sNT representations might deviate from their local objectives, even if we have only one objective (e.g. at the last node in the  network). Therefore, the main idea behind our learning approach is to intoduce a local propagation constraints in order to compensate and correct this deviation, but, in a localized manner using  the sNT representations, a  \textit{corrective sparsifying nonlinear transform} (c-sNT) representations and representations that exactly satisfy the local goal from the current node and the closely connected nodes. 


\textbf{{Corrective Sparsifiying Nonlinear Transform}} Assume that 
a correction vector $\boldsymbol{\nu}_{l, \{ c,k\}} \in \Re^{M_{l}}$ and a threshold parameter $\lambda_{l, 1} \in \Re_+$ are given.
Denote ${\bf b}_{l, \{c,k\}}={\bf q}_{l, \{c,k\}}-\boldsymbol{\nu}_{l, \{ c,k\}}$, 
then similarly to the sNT, we define the c-sNT representation ${\bf y}_{l,\{c,k\}}$ at level $l$ as: 
\vspace{-.05in}
\begin{align}
\! \! \! \! {\bf y}_{l,\{c,k\}}= {\rm sign}( {\bf b}_{l, \{c,k\}}) \odot \max( \vert {\bf b}_{l, \{c,k\}} \vert-\lambda_{l,1}{\bf 1}, {\bf 0}). \label{eq:nonlinear:representation}
\end{align}
The 
c-sNT \eqref{eq:nonlinear:representation} at node level $l$ is defined on top of the 
sNT representations \eqref{eq:sparse:representation} at node level $l-1$. 
We 
point out that the vector $\boldsymbol{\nu}_{l, \{ c,k\}}={\bf p}_{l, \{ c,k\}}+{\bf t}_{l, \{ c,k\}}$ is a linear composition of two components. The component ${\bf p}_{l, \{ c,k\}}$ has a deviation corrective role that comes from the local propagation constraint, whereas the component ${\bf t}_{l, \{ c,k\}}$ has a goal related role, which enables the local goal to be satisfied. 
We also refer to $\boldsymbol{\nu}_{l, \{ c,k\}}$ as the portion of the parameter set $\mathcal{P}_{l, \{c,k \}}={\{ {\bf A}_{l-1},  \boldsymbol{\nu}_{l, \{ c,k\}}, \lambda_{l,1} \} \}}$ that describes the c-sNT. 
  
During learning, in contrast to the predefined "static" activation function, the c-sNT \eqref{eq:nonlinear:representation}  representations are dynamically estimated 
depending on the deviation of the sNT representations from the local goal and the local propagation constraints, which is essential in order to add a corrective element in the update of the NN weights. More importantly, 
the c-sNTs 
play a crucial role in enabling decoupled and parallel updates of the weights in the NN.   
\vspace{-0.04in}


\subsection{Contributions}
\vspace{-0.06in}
In the following, we summarize our contributions.

(i) We introduce a learning problem formulation, 
which to the best of our knowledge is first of this kind that:
\vspace{-0.1in}
\vspace{-0.01in}
\begin{enumerate}
\item[(a)] explicitly addresses a trade-off between (i) satisfying local objectives at each node related to the corresponding local representation goal and (ii) achieving desired data propagation through the network nodes that enables attaining a targeted 
representations at the last node in the network under (iii) local propagation constraints
\vspace{-0.1in}

\item[(b)] offers the possibility of posing a wide class of arbitrary local goals and propagation constraints while enabling efficient estimation of the sNT, c-sNT and the network weights

\vspace{-0.1in}
\item[(c)]  provides interpretation of the local learning dynamics by connecting it to a \textit{local diffusion model} \citet{Kittel:Charles_Kroemer:Herbert_1980} or change of the local flow.
\end{enumerate}
(ii) We propose a novel learning strategies that can operate in synchronous or asynchronous mode, which we implement by an efficient algorithm with parallel execution that iterates between two stages: 

\vspace{-0.01in}
\begin{enumerate}
\vspace{-0.1in}
\item[1)] estimation of the sNT representations and the exact goal satisfying representations and
\vspace{-0.1in}
\item[2)] estimation of the c-sNT representations and the actual network weights.
\end{enumerate}
\vspace{-0.1in}
(iii) We provide theoretical analysis and empirically validate the potential of our approach. 
Our results demonstrate that the proposed learning principle allows targeted representations to be attained w.r.t. a goal set only at one node located anywhere in the NN.


\vspace{-0.1in}
\subsection{Notations}
\vspace{-0.0in}
A variable at node level $l$ has a subscript $*_l$. Scalars, vectors and matrices are denoted by usual, bold lower and bold upper case symbols as ${x}_l$, ${\bf x}_l$ and ${\bf X}_l$. A set of data samples from $C$ classes is denoted as ${\bf Y}_l=[{\bf Y}_{l,1},...,{\bf Y}_{l,C}] \in \Re^{M_l \times CK}$. Every class $c \in \{1,...,C\}$ has $K$ samples, ${\bf Y}_{l,c}$$=[{\bf y}_{l, \{c,1\} },...,{\bf y}_{l, \{c,K\}}] \in \Re^{M_l \times K}$. We denote the $k-$th representation from class $c$ at level $l$ as ${\bf y}_{l, \{c,k\}} \in \Re^{{M}_l}$, $\forall c \in \{1,...,C\}$, $\forall k \in \{1,...,K\}$, $\forall l \in \{1,...,L\}$.
The $\ell_p-$norm, nuclear norm, matrix trace and Hadamard product are denoted as $\Vert.\Vert_p$, $\Vert. \Vert_{*}$, $Tr()$ and $\odot$, respectively. 
The first order derivative of a function $\mathcal{L}({\bf Y}_{l})$ w.r.t. ${\bf Y}_l$ is denoted as $\frac{\partial \mathcal{L}({\bf Y}_{l})}{\partial {\bf Y}_l}$. We denote $\vert {\bf y}_{l, \{ c,k \}} \vert$ as the vector having as elements the absolute values of the corresponding elements in ${\bf y}_{l, \{ c,k \}}$. 

\vspace{-0.1in}

\vspace{-0.0in}
\section{The Learning Problem }
\label{sec:The:learning:algorithm}


In this section, we present our problem formulation, explain the local goal and the local propagation constraint and unveil 
our learning target. 

We take into account one extended version of feed-forward NN. At each network node $l$, our learning concept 
considers c-sNT representations ${\bf Y}_l$, sNT representations ${\bf U}_l$, representations ${\bf G}_l \in \Re^{M_l \times CK}$ that exactly satisfy the local goal, local objectives related to the corresponding desired representations and a local propagation constraints. In the most general form, we introduce our problem formulation in the following form:
\vspace{-0.12in}
\begin{equation}
\begin{aligned}
\hat{\boldsymbol{\Omega} }= \min_{\boldsymbol{\Omega} } \sum_{l=1}^L & \left( \underbrace{ \mathcal{R}_{1}(l) }_{\icol{\text{NT transform} \\ \text{errors}} }\right. \!\!+\!\!\underbrace{\mathcal{R}_{2}(l)}_{\icol{\text{weight} \\ \text{constraints}}}  + \underbrace{\mathcal{A}(l)}_{\icol{\text{sparsity} \\ \text{constraints}}}+ \\ 
&\hspace{.3in} \underbrace{\mathcal{R}_3(l) }_{\icol{\text{local goal } \\ \text{constraint}}} \!\!\!+
\left. \underbrace{\mathcal{R}_{4}(l)}_{\icol{\text{local propagation } \\ \text{constraint} }} \right), \label{eq:global:problem:formulation}  \\
\hspace{.0in} \text{ subject to}&    \text{ ${ }$ } \text{ ${ }$ } \mathcal{U}(\!\!\!\underbrace{{\bf G}_l}_{\icol{ \text{exact goal}\\ \text{stisfying represetaton}}} \!\!\!)=0, \forall l, 
\end{aligned}
\end{equation} 
where $\boldsymbol{\Omega}=\{ {\bf A}_0,..., {\bf A}_{L-1}, {\bf U}_1,...,{\bf U}_{L},$ $ {\bf Y}_1,.,{\bf Y}_{L},{\bf G}_1,...,{\bf G}_{L}, {\bf B}_0,..., {\bf B}_{L-1} \}$ are the network parameters used during learning. We denote ${\bf A}_l$ as a forward weight, whereas ${\bf B}_{l}\in \Re^{M_{l} \times M_{l+1}}$ as a backward weight. Note that the actual network weighs used during testing are $\{ {\bf A}_0,..., {\bf A}_{L-1} \}$ and the resulting network representation from consecutively using our sNT\footnote{
The sNTs are one type of nonlinear activation function \eqref{eq:background:solution:1}. } are $\{ {\bf U}_1,..., {\bf U}_{L} \}$. 
In the following subsections, we define and explain each of the components in our problem formulation.

\subsection{Nonlinear Transform Errors and Error Vectors}

The term $\mathcal{R}_1(l)=\mathcal{L}({\bf A}_{l-1}{\bf U}_{l-1}, {\bf Y}_{l})+\mathcal{L}({\bf B}_{l}{\bf U}_{l+1}, {\bf Y}_{l})+$ $\mathcal{L}({\bf A}_{l-1}{\bf U}_{l-1}, {\bf G}_{l})$ 
models three nonlinear transform errors at node level $l$. The first two are related to the sNT representations ${\bf U}_{l-1}$, ${\bf U}_{l+1}$ and the c-sNT ${\bf Y}_l$ representations  and the last one is related  
to the ${\bf G}_l$ representations. Term $\mathcal{L}({\bf A}_{l-1}{\bf U}_{l-1}, {\bf Y}_{l})=\frac{1}{2}\Vert {\bf A}_{l-1}{\bf U}_{l-1}- {\bf Y}_{l} \Vert_F^2$ measures the deviations of the gNT representations away from the linear transform representations ${\bf Q}_l={\bf A}_{l-1}{\bf U}_{l-1}$, whereas:
\begin{equation}
\begin{aligned}
\textit{te:}&\frac{\partial \mathcal{L}({\bf Q}_{l},{\bf Y}_{l})}{\partial {\bf Y}_{l}}=&& {\bf Y}_{l}-{\bf Q}_{l}, 
\label{eq:error:vectors:nte}
\end{aligned}
\end{equation}
represent the corresponding deviation vectors. Also, the term 
$\mathcal{L}({\bf A}_{l-1}{\bf U}_{l-1}, {\bf G}_{l})=\frac{1}{2}\Vert {\bf A}_{l-1}{\bf U}_{l-1}- {\bf G}_{l} \Vert_F^2$ has similar role, related to ${\bf G}_l$, respectively. 

In addition, $\mathcal{L}({\bf A}_{l-1}{\bf U}_{l-1}, {\bf Y}_{l})$ is related to the forward propagation of ${\bf U}_{l-1}$ through ${\bf A}_{l-1}$, whereas $\mathcal{L}({\bf B}_{l}{\bf U}_{l+1}, {\bf Y}_{l})$ is related to the backward propagation of ${\bf U}_{l+1}$ through ${\bf B}_{l}$. 
We introduce 
the backward weights to enable regularization of the local propagation in a localized manner. 
We can also model ${\bf B}_l={\bf A}^T_l$, but, in order to present the full potential of our approach, we consider that ${\bf B}_l$ is different from ${\bf A}_l^T$. 

\subsection{Weights Constraint}
The term $\mathcal{R}_2(l)=\mathcal{V}({\bf A}_{l-1})+\mathcal{L}({\bf A}_{l}, {\bf B}_{l})$ models the properties of the weights that connect nodes at levels $l-1$ and $l$, as well as nodes at levels $l$ and $l+1$, where $\mathcal{V}({\bf A}_{l-1})=\frac{\lambda_{l,2}}{2}\Vert {\bf A}_{l-1} \Vert_{F}^2+\frac{\lambda_{l,3}}{2}\Vert{\bf A}_{l-1}{\bf A}_{l-1}^T-{\bf I} \Vert_F^2-\lambda_{l,4}\log \vert \det {\bf A}_{l-1}^T{\bf A}_{l-1} \vert$ and $\mathcal{L}({\bf A}_{l}, {\bf B}_{l})=\frac{\lambda_{l,5}}{2}\Vert {\bf A}_{l}-{\bf B}_{l}^T \Vert_F^2$ 
are used to regularize the conditioning, the coherence 
of ${\bf A}_{l}$ \citet{Kostadinov2018:EUVIP}, and the similarity between ${\bf A}_{l}$ and ${\bf B}_{l}^T$, respectively.

\vspace{-.06in}
\subsection{Sparsity Constraints}
Our sparsity constraint is defined on the sNT representations ${\bf U}_{l}=[{\bf u}_{l, \{1,1 \}},...,{\bf u}_{l, \{C,K \}}]$, the c-sNT representations ${\bf Y}_{l}=[{\bf y}_{l, \{1,1 \}},...,{\bf y}_{l, \{C,K \}}]$ and the representations ${\bf G}_{l}=[{\bf g}_{l, \{1,1 \}},...,{\bf g}_{l, \{C,K \}}]$ that exactly satisfy the specified local goal as $\mathcal{A}(l)=\lambda_{l,1}\sum_{c=1}^{C}\sum_{k=1}^{K} \left( \Vert {\bf u}_{l,\{c,k\}} \Vert_1+\Vert {\bf y}_{l,\{c,k\}} \Vert_1+\Vert {\bf g}_{l,\{c,k\}} \Vert_1\right)$. 

\subsection{Local Goal Constraint}
\label{LocalPropagatioModel:DynamicandInterpretations}




Before defining the term $\mathcal{R}_3(l)$, we first define our local goal that is explicitly set on ${\bf G}_{l}=[{\bf g}_{l, \{1,1 \}},...,{\bf g}_{l, \{C,K \}}]$. That is, knowing the corresponding labels, 
we express it in a form of a discrimination constraint, which is defined as $\mathcal{U}({\bf G}_{l})=\lambda_{l,0}D({\bf G}_{l})=$  $\lambda_{l,0}\sum_{\icol{ c1, c1 \neq c} }\sum_{k1}( \Vert {\bf g}^{+}_{l,\{c,k\}}\odot{\bf g}^{+}_{l, \{ c1,k1\}} \Vert_1+$ $\Vert {\bf g}^{-}_{l, \{c,k \}}\odot{\bf g}^{-}_{l, \{c1,k1\}}\Vert_1$$+\Vert {\bf g}_{l,\{c,k\}}\odot{\bf g}_{l, \{ c1,k1\}} \Vert_2^2 )$, where ${\bf g}_{l, \{c,k\}}={\bf g}_{l, \{ c,k \}}^+-{\bf g}_{l, \{ c,k \}}^-$, ${{\bf g}_{l, \{c1,k1\} }^+=\max({\bf g}_{l, \{ c1, k1 \}}, {\bf 0})}$ and ${{\bf g}_{l, \{ c1,k1\}}^-=\max(-{\bf g}_{l ,\{ c1,k1\}}, {\bf 0})}$ 
\citet{Kostadinov:ICLR2018}. 

By considering the representations ${\bf G}_{l}$ and 
${\bf U}_{l}$, 
beside the \textit{te} vectors, we also define the local goal error (\textit{ge}) vectors as: 
\vspace{-0.05in}
\begin{equation}
\begin{aligned}
\textit{ge:}&\frac{\partial \mathcal{G}({\bf G}_{l},{\bf U}_{l})}{\partial {\bf U}_{l}}=&& {\bf U}_{l}-{\bf G}_{l}, 
\label{eq:error:vectors}
\end{aligned}
\end{equation}
which represent the deviation of the  representations ${\bf U}_{l}$ 
away from the representations ${\bf G}_{l}$. 

We would like ${\bf U}_{l}$ to match ${\bf G}_{l}$, but in order to have more freedom in modeling a wide range of goals as well as allow decoupled update per the network weights, we do not set it as explicit constraint on ${\bf U}_{l}$. Rather, we define it as follows:
 \vspace{-0.066in}
\begin{equation}
\begin{aligned}
\hspace{-0.21in} &\mathcal{R}_3(l)=\sum_{\icol{c=1}}^{C}\sum_{\icol{k=1}}^{K}\psi( r_g(c,k) ), \text{${}$ } \text{${}$ } \text{${}$ } \text{${}$ } \text{${}$ } \text{${}$ } \text{${}$ } \text{${}$ } {r}_g(c,k)=\\
\hspace{-0.21in}&\left(\frac{\partial \mathcal{L}({\bf q}_{l, \{c,k \}},{\bf y}_{l, \{c,k \}})}{\partial {\bf y}_{l, \{c,k \}}}\right)^T \frac{\partial \mathcal{L}({\bf g}_{l, \{c,k \}},{\bf u}_{l, \{c,k \}})}{\partial {\bf u}_{l, \{c,k \}}},
\end{aligned}
\label{eq:global:problem:error:goal:constraint}
\end{equation}
where in the simplest case we let $\psi( r_g(c,k) )=r_g(c,k) $.  



\subsection{Local Propagation Constraint}
Our local propagation constraint takes into account the deviation vectors \eqref{eq:error:vectors:nte} and 
\eqref{eq:error:vectors} that where explained in the previous two subsection and has a 
diffusion \citet{spivak1980calculus} and 
\citet{Kittel:Charles_Kroemer:Herbert_1980} related form that we define as:
 \vspace{-0.066in}
\begin{equation}
\begin{aligned}
\hspace{-0.22in} &\mathcal{R}_4(l)=\sum_{\icol{c=1}}^{C}\sum_{\icol{k=1}}^{K}\psi( r_p(c,k) ), \text{${}$ } \text{${}$ } \text{${}$ } \text{${}$ } \text{${}$ } \text{${}$ } \text{${}$ } \text{${}$ } {r}_p(c,k)=\\
\hspace{-0.22in}&\left(\frac{\partial \mathcal{L}({\bf q}_{l, \{c,k \}},{\bf y}_{l, \{c,k \}})}{\partial {\bf y}_{l, \{c,k \}}}\right)^T  \! \! \! \nabla^2 \mathcal{G}({\bf u}_{l-1, \{c,k \}}, {\bf u}_{l+1, \{c,k \}} ),
\end{aligned}
\label{eq:global:problem:error:propagation:flow}
\end{equation}
\vspace{-0.1in}
{\flushleft where} ${\nabla^2 \mathcal{G}({\bf U}_{l-1}, {\bf U}_{l+1} ) }=\left[ \lambda_{l,f}{\bf B}_{l}\frac{\partial \mathcal{G}({\bf G}_{l+1},{\bf U}_{l+1})}{\partial {\bf U}_{l+1}}\right.+ \left.  \lambda_{l, b}{\bf A}_{l-1}\frac{\partial \mathcal{G}({\bf G}_{l-1},{\bf U}_{l-1})}{\partial {\bf U}_{l-1}} \right]={\bf F}_{l, f}+{\bf F}_{l, b} $ is the local diffusion term, 
representing the vectors for the \textit{change of the local propagation flow}, and $\lambda_{l,b}$ and $\lambda_{l,f}$ are regularization parameters. 
Term ${\nabla^2 \mathcal{G}({\bf U}_{l-1}, {\bf U}_{l+1} ) }$ compactly describes 
the deviations of the representation ${\bf Y}_l$ at node level $l$ w.r.t. the propagated \textit{ge} vectors 
$\frac{\partial \mathcal{G}({\bf G}_{l-1},{\bf U}_{l-1})}{\partial {\bf U}_{l-1}}$ and $\frac{\partial \mathcal{G}({\bf G}_{l+1},{\bf U}_{l+1})}{\partial {\bf U}_{l+1}}$
from node levels $l-1$ and $l+1$, 
through ${\bf A}_{l-1}$ and ${\bf B}_{l}$, respectively \footnote{Note that when there is no local goals defined at node levels $l-1$ and $l+1$, the representations ${\bf G}_{l-1}$ and ${\bf G}_{l+1}$ are zero vectors, 
\eqref{eq:global:problem:error:propagation:flow} 
regularizes the \textit{local propagation flow} and 
takes the form as 
$\mathcal{R}_{3}(l)= \sum \psi( (\frac{\partial \mathcal{L}({\bf q}_{l, \{c,k \}},{\bf y}_{l, \{c,k\}})}{\partial {\bf y}_{l, \{c,k \}}})^T  \nabla \mathcal{G}({\bf u}_{l-1, \{c,k \}}, {\bf u}_{l+1, \{c,k \}}) ) $, where ${\nabla \mathcal{G}({\bf U}_{l-1}, {\bf U}_{l+1} ) }=\left[ \lambda_{l,f}{\bf B}_{l}{\bf U}_{l+1} \right.+ \left.  \lambda_{l, b}{\bf A}_{l-1}{\bf U}_{l-1} \right] $.  }.

\subsubsection{Local Propagation Dynamics}  
\label{LocalPropagationDynamics} To explain the local propagation dynamics, we analyze its influence in the learning problem. 
The terms 
$Tr \{\frac{\partial \mathcal{L}({\bf Q}_{l},{\bf Y}_{l})}{\partial {\bf Y}_l} {\bf F}_{l,f} \}$ and $Tr \{\frac{\partial \mathcal{L}({\bf Q}_{l},{\bf Y}_{l})}{\partial {\bf Y}_l} {\bf F}_{l,b} \}$ 
will be zero if the \textit{ge} 
\eqref{eq:error:vectors} 
or the \textit{te} vectors 
\eqref{eq:error:vectors:nte} are zero. In that case, ether the forward or backward local propagation constraint is totally satisfied, since a sparse version of ${\bf Q}_{l-1}$ (or ${\bf A}_{l}{\bf U}_{l}$) 
equals the representations ${\bf G}_{l-1}$ (or ${\bf G}_{l+1}$), ether  ${\bf Q}_{l}$ equals\footnote{In general ${\bf A}_{l-1}{\bf U}_{l-1}$ is not sparse. However, it is possible 
${\bf A}_{l}{\bf U}_{l-1}$ to have any desirable properties within a very small error. 
} to the representations ${\bf Y}_{l}$. 
While when the affine combination between the propagated \textit{ge} 
vectors \eqref{eq:error:vectors} from node levels $l-1$ and $l+1$, through ${\bf A}_{l-1}$ and ${\bf B}_{l}$ 
are orthogonal to the \textit{te} vectors 
\eqref{eq:error:vectors:nte} at level $l$, 
\textit{i.e.}, 
$Tr \{ {\nabla^2 \mathcal{G}({\bf U}_{l-1}, {\bf U}_{l+1} ) } \frac{\partial \mathcal{L}({\bf Q}_l,{\bf Y}_{l})}{\partial {\bf Y}_l}^T \}=0$ then the \textit{change of the local propagation flow is preserved} w.r.t. the \textit{te} vectors $\frac{\partial \mathcal{L}({\bf Q}_l,{\bf Y}_{l})}{\partial {\bf Y}_l}$. In that case, an alignment is achieved between the
sNT representations and the c-sNT representations regardless of whether the local goal is achieved $Tr \{ \frac{\partial \mathcal{L}({\bf Q}_l,{\bf Y}_{l})}{\partial {\bf Y}_l}^T\frac{\partial \mathcal{L}({\bf G}_l,{\bf U}_{l})}{\partial {\bf U}_l} \}=0$ or not $Tr \{ \frac{\partial \mathcal{L}({\bf Q}_l,{\bf Y}_{l})}{\partial {\bf Y}_l}^T\frac{\partial \mathcal{L}({\bf G}_l,{\bf U}_{l})}{\partial {\bf U}_l} \} \neq 0$. 

\begin{figure}[t]
\centering
\begin{center}
\begin{minipage}[b]{1\linewidth}
\centering
\centerline{\includegraphics[width=\columnwidth]{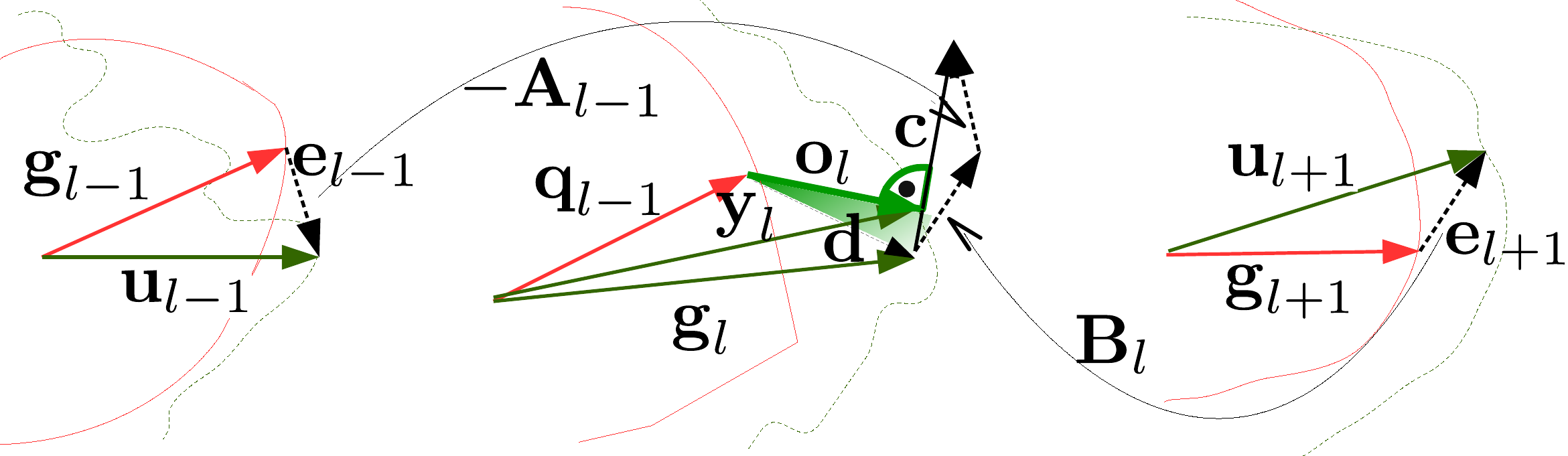}}
\vskip 0.01in
\begin{tabular}{c@{\hspace{60pt}}c@{\hspace{60pt}}c}
$l-1$ & $l$& $l+1$ 
\end{tabular}
\end{minipage}
\end{center}
\vskip -0.0in
\caption{The illustration of the learning dynamics that we explained  in detain in \ref{LocalPropagationDynamics} 
and \ref{Trade-OffsandLearningTarget}. The red and green curves represent the space were the nonlinear transform and the desired representations live, respectively. The goal error vectors at node levels $l-1$ and $l+1$ are ${\bf e}_{l-1}={\bf u}_{l-1}-{\bf g}_{l-1}$ and ${\bf e}_{l+1}={\bf u}_{l+1}-{\bf g}_{l+1}$. The change of the local propagation flow is denoted as ${\bf c} \simeq {\bf B}_{l}{\bf e}_{l+1}+{\bf A}_{l-1}{\bf e}_{l-1}$, the c-sNT error vector is ${{\bf o}_l={{\bf A}_{l-1}{\bf u}_{l-1}-{\bf y}_{l}}}$ and ${\bf c}^T{\bf o}_l$ is the local propagation term \eqref{eq:global:problem:error:propagation:flow}. The set of directions colored in green highlights a trade-off that we described in \ref{Interpretation} and \ref{Trade-OffsandLearningTarget}.}
\label{intro:image:prior:dynamics}
\vskip -0.0019in
\end{figure}

\label{Interpretation}
To understand what exactly this alignment means w.r.t. the updates in the network weighs, the c-sNTs and the sNTs, we analyze one commonly used network as an example. Assume that we have a feed-forward network with local propagation constraints and no local goal constraints at all node levels $l$, except one goal at the last network node.  
Lets say that for node level $l$, $\frac{\partial \mathcal{L}({\bf Q}_l,{\bf Y}_{l})}{\partial {\bf Y}_l}$
are orthogonal to ${\nabla^2 \mathcal{G}({\bf U}_{l-1}, {\bf U}_{l+1} ) }$. This means that the 
used weighs ${\bf A}_{l-1}$ and ${\bf B}_{l}$ 
are not adding additional deviation in the change of the local propagation flow ${\nabla^2 \mathcal{G}({\bf U}_{l-1}, {\bf U}_{l+1} ) }$ 
when 
we propagate the representations ${\bf U}_{l-1}$ forward through ${\bf A}_{l-1}$. 
In that case, the achieved alignment 
w.r.t. the preserved local propagation flow indicates that the c-sNT representations do not contain any additional components different then the sNT representations. Therefore,  the c-sNT representations do not add additional "information", which can be used in the update of ${\bf A}_{l-1}$ or ${\bf B}_l$ to further reduce the term \eqref{eq:global:problem:error:propagation:flow}. 
Since when $\frac{\partial \mathcal{L}({\bf Q}_l,{\bf Y}_{l})}{\partial {\bf Y}_l}$
are orthogonal to ${\nabla^2 \mathcal{G}({\bf U}_{l-1}, {\bf U}_{l+1} ) }$, \eqref{eq:global:problem:error:propagation:flow} is already zero. 

Lets say that the local propagation flow is preserved at every network node. In addition, lets say that at the last network node the local goal is also satisfied. Then, by propagating the data forward through the network using the sNT, the network weights ${\bf A}_l$ are estimated such that they will not add additional deviation in the consecutive estimation of the sNT representations. 
Thereby, this will allow to attain the desired representations at the last network node. 
Otherwise, if there are deviations, 
the corresponding components from term \eqref{eq:global:problem:error:propagation:flow} 
should be used to 
add a locally adjusted correction element 
in the update of the weighs and thus to enforce preservation in the change of the propagation flow. 

\subsubsection{Trade-Offs and Learning Target} 
\label{Trade-OffsandLearningTarget} If a local goal constraint \eqref{eq:global:problem:error:goal:constraint} is included, satisfying its objective adds additional deviation. Therefore, at one network node, we have a trade-off between satisfying a local goal and a local propagation constraint \eqref{eq:global:problem:error:propagation:flow}
while over all NN nodes 
we have a trade-off between (i) satisfying a local goal, (ii) satisfying a local propagation constraint and (iii) achieving desired data propagation through the NN that enables attaining targeted representations at the last NN node.

In relation to the network parameters $\mathcal{P}=\{\mathcal{P}_1,...,\mathcal{P}_L \}$ and $\mathcal{S}=\{ \mathcal{S}_{1}, ..., \mathcal{S}_{L}\}$ that are used for training, our learning problem  \eqref{eq:global:problem:formulation} targets to estimate the parameter set $\mathcal{S}$ for the sNTs that approximate the parameter set $\mathcal{P}$ of the c-sNTs. One sNT that is defined by $\mathcal{S}_l=\{ {\bf A}_l, \tau_{l} \}$ approximates one set of c-sNTs that is defined by $\mathcal{P}_l=\{ \mathcal{P}_{l, \{1,1\}},..., \mathcal{P}_{l, \{C,K\}}\} \}$.  That is, for every node at level $l$, given ${\tau_l}$, we would like to estimate ${\bf A}_l$ such that 
the c-sNT representations 
${\bf Y}_{l}$ 
become 
equal to the 
sNT representations ${\bf U}_{l}$ while our local goals and local propagation constraints are satisfied. 
An illustration of the learning dynamics as well as the involved trade-offs is given in Figure \ref{intro:image:prior:dynamics}.

\section{The Learning Strategy}




This section presents 
the solution to \eqref{eq:global:problem:formulation} in synchronous and asynchronous scheduling regime by essentially 
using two variants of one 
learning principle. 
\subsection{The Learning Algorithm} 
\label{TheLearningAlgorithm}
Our learning algorithm iteratively updates the network parameters in two stages. Stage one updates the sNT representations ${\bf U}_l$ and the exact goal satisfying representations ${\bf G}_l$ while stage two estimates the c-sNT representations ${\bf Y}_l$ and the weights ${\bf A}_{l-1}$ and ${\bf B}_l$.
\subsubsection{Stage One}

Given the weighs ${\bf A}_{l}$, this stage computes ${\bf U}_l$ and ${\bf G}_l$.



\textbf{Estimating ${\bf U}_l$ Approximatively by Discarding Constraints and Coupling } In this stage, we let ${\bf U}_l={\bf Y}_l$, fix all the variables in problem \eqref{eq:global:problem:formulation} except ${\bf U}_l$, disregard the local goal, the local propagation constraint 
and the coupling over two representations at levels $l-1$ and $l$, then  per node level $l$, problem \eqref{eq:global:problem:formulation} reduces to:
\begin{equation}
\begin{aligned}
\hat{{\bf U}}_{l}=\min_{{\bf U}_{l}}   \mathcal{L} ( {\bf Q}_l,{\bf U}_{l} )+\lambda_{l,1}\sum_{c=1}^C\sum_{k=1}^K\Vert  {\bf u}_{l, \{c,k\}} \Vert_1,
\end{aligned}
\label{eq:global:problem:formulation:reduction:goal}
\end{equation}
where the solution per single $\hat{{\bf u}}_{l, \{c,k\}}$ is exactly the  sNT \eqref{eq:sparse:representation}. Therefore, computing ${\bf U}_l$ by propagating forward 
 through the network with consecutive execution of the sNT, is in fact an approximative solution w.r.t.  \eqref{eq:global:problem:formulation}. 

\textbf{Estimating ${\bf G}_l$ Approximatively by Discarding Local Goal and Local Propagation Constants} Given ${\bf Q}_l={\bf A}_{l-1}{\bf U}_{l-1}$, 
if we disregard the local propagation constraint and the local goal constraint, per node level $l$, ${\bf G}_{l}$ are defined as the solution of an optimization problem where ${\bf G}_l$ has to be close to the linear transform representations ${\bf Q}_l={\bf A}_{l-1}{\bf U}_{l-1}$ under the sparsity constraint and the discrimination\footnote{In general, one might model different goals for the representations ${\bf G}_{l}$ by 
defining a corresponding function ${\mathcal{U}({\bf G}_l)}$.} constraint,\textit{ i.e.}:
\begin{equation}
\begin{aligned}
\! \! \!  \hat{{\bf G}}_{l} \!  = \!  \min_{{\bf G}_{l}}   &\mathcal{L} ( {\bf Q}_l,{\bf G}_{l} )\! +\! \lambda_{l,1}\sum_{c=1}^C\sum_{k=1}^K\Vert  {\bf g}_{l, \{c,k\}} \Vert_1, \!\\ 
&\text{subject to } {\mathcal{U}}({\bf G}_{l})=0.
\end{aligned}
\label{eq:global:problem:formulation:reduction:goal}
\end{equation}
In \textit{Appendix} A, we give an iterative solution to \eqref{eq:global:problem:formulation:reduction:goal} with closed form updates at the iterative steps. 


\begin{figure}[t]
\centering
\begin{center}
\begin{minipage}[b]{0.9\linewidth}
\centering
\centerline{\includegraphics[width=\columnwidth]{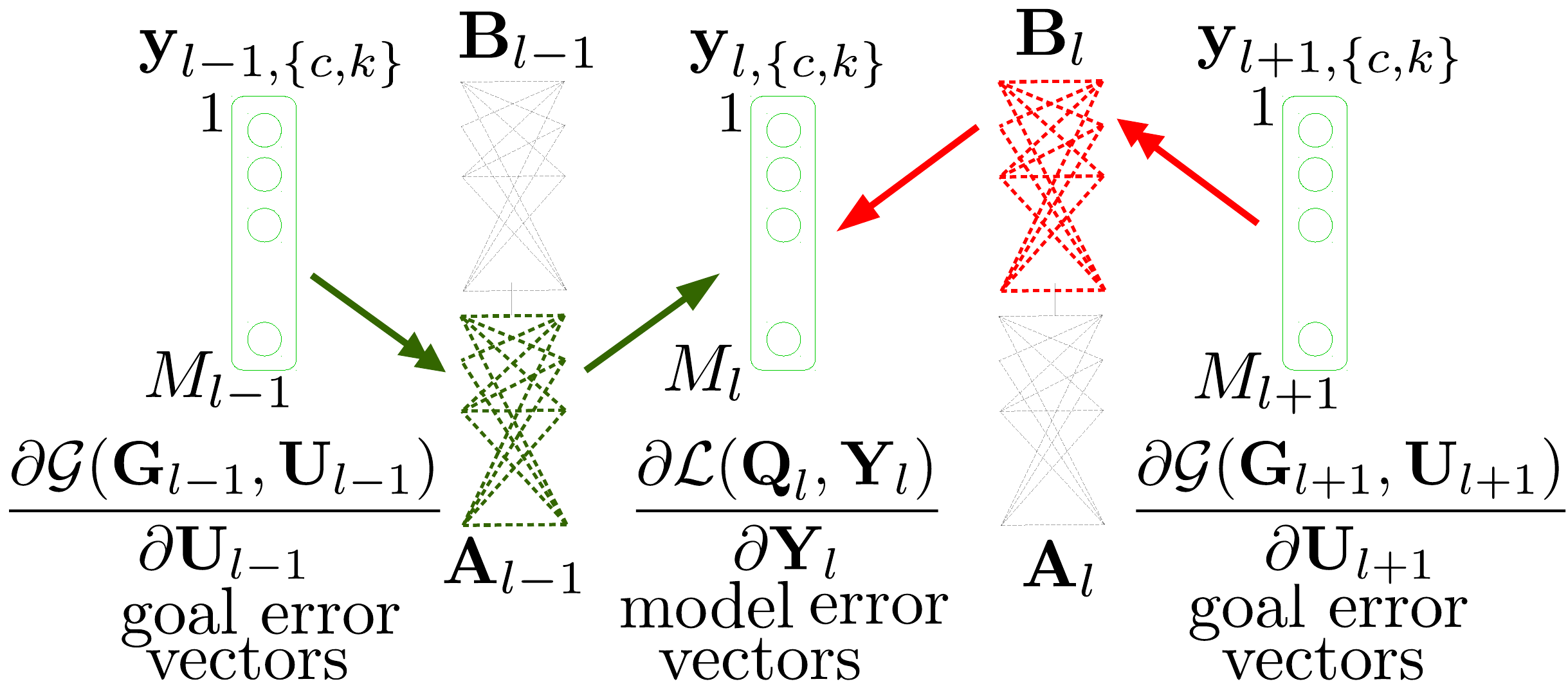}}
\begin{tabular}{c@{\hspace{60pt}}c@{\hspace{60pt}}c}
$l-1$ & $l$& $l+1$ 
\end{tabular}
\end{minipage}
\end{center}
\vskip -0.19in
\caption{The resulting local network when fixing all the network variables except ${\bf A}_{l-1}, {\bf B}_{l}$ and ${\bf Y}_l$, where ${\bf Y}_l$ are the c-sNT representations, and ${\bf U}_{{l-1}}$ and ${\bf U}_{{l+1}}$ are the sNT representations. }
\vskip -0.0019in
\label{local.Network}
\end{figure}

\subsubsection{Stage Two}
Given all of the currently estimated ${\bf U}_{l}$ and ${\bf G}_{l}$, note that \eqref{eq:global:problem:formulation} decomposes over subproblems that are separable per every parameter subset $\boldsymbol{\varsigma}_l=\{ {\bf Y}_l, {\bf A}_{l-1}, {\bf B}_{l} \}$, 
(
Figure \ref{local.Network}). This allows parallel update on all subsets $\varsigma_l$ of network parameters, since 
the parameter set $\boldsymbol{\varsigma}_{l_1}$ does not share parameters with any other $\boldsymbol{\varsigma}_{l_2}$, \textit{i.e.}, $\boldsymbol{\varsigma}_{l_1} \bigcap \boldsymbol{\varsigma}_{l_2}=\emptyset, \forall l_1 \neq l_2$. 
The learning subproblems per the decoupled sets $\boldsymbol{\varsigma}_l$ have one common form. In the following, we present it and give the solution. 

Let all the variables in \eqref{eq:global:problem:formulation} be fixed except $\boldsymbol{\varsigma}_l=\{ {{\bf Y}}_l, {{\bf A}}_{l-1}, {{\bf B}}_{l} \}$, 
then \eqref{eq:global:problem:formulation} 
reduces to the following problem: 
\begin{equation}
\{ \hat{{\bf Y}}_l, \hat{{\bf A}}_{l-1}, \hat{{\bf B}}_{l} \}=\min_{ \{ {{\bf Y}}_l, {{\bf A}}_{l-1}, {{\bf B}}_{l} \} }\sum_{j=1}^4{\mathcal{R}}_{j}(l)+{\mathcal{A}}(l).
\label{eq:global:problem:formulation:reduction.0-1}
\end{equation}
\textbf{Estimating $\boldsymbol{\varsigma}_l=\{{\bf Y}_{l}, {\bf A}_{l-1}, {\bf B}_{l}\}$ Exactly} Problem \eqref{eq:global:problem:formulation:reduction.0-1} is still non-convex. Nevertheless, to solve \eqref{eq:global:problem:formulation:reduction.0-1}, we propose an alternating block coordinate descend algorithm, where we iteratively update one variable from the set of variables $\varsigma_l=\{ {\bf Y}_l, {\bf A}_{l-1}, {\bf B}_{l} \}$ while keeping the rest fixed. 
It has three steps: (i) 
estimation of the c-sNT representation ${\bf Y}_l$, 
(ii) estimation of the forward weights ${\bf A}_{l-1}$ and (iii) estimation of the backward weights  ${\bf B}_{l}$. In the following, we explain the steps of the proposed solution. 

\textit{\underline{$-$ c-sNT Representation Estimation}} 
\label{label:sec:representation:estimate}
Let all the variables in problem \eqref{eq:global:problem:formulation:reduction.0-1} be given except ${\bf y}_{l, \{c,k\}}$ then \eqref{eq:global:problem:formulation:reduction.0-1} 
reduces to the following \textit{constrained projection problem}:
\begin{align}
&\hat{\bf y}=\arg \min_{ \icol{{\bf y} } }\frac{1}{2}\Vert {\bf q}-{\bf y} \Vert_2^2+{\boldsymbol{\nu}^T}{\bf y}+\lambda_{l,1}{\bf 1}^T\vert {\bf y} \vert, \label{eq:global:problem:formulation:reduction:representation:estimation} 
\end{align}
where:
\begin{align}
&{\bf y}={\bf y}_{l, \{c,k\}}, \nonumber \\
&{\bf q}={\bf A}_{l-1}{\bf u}_{l-1,\{c,k\}},\nonumber \\
&\boldsymbol{\nu}={\bf p}_{l, \{c,k\}}+{\bf t}_{l, \{c,k\}}, \nonumber \\
&{\bf t}_{l, \{c,k\}}=\frac{\partial \mathcal{L}({\bf g}_{l, \{c,k \}},{\bf u}_{l, \{c,k \}})}{\partial {\bf u}_{l, \{c,k \}}}, \label{eq:global:problem:formulation:reduction.0-1-1}  \\
&{\bf p}_{l, \{c,k\}}=\nabla^2 \mathcal{G}({\bf u}_{l-1, \{c,k \}}, {\bf u}_{l+1, \{c,k \}} ), \nonumber  
\end{align}
and it has a closed form solution which exactly matches the expression for the c-sNT \eqref{eq:nonlinear:representation}. 
The proof is given in \textit{Appendix} B. In addition, note that by \eqref{eq:nonlinear:representation} all ${\bf y}_{l, \{c,k\}}$ at node level $l$ can be computed in parallel. 

The empirical expectation 
of ${\mathbb E}[{\boldsymbol{\nu}^T}{\bf y}]$ 
induced by the local goal constraint and the local propagation constraint can also be considered as the {empirical risk} for the sNT \eqref{eq:sparse:representation}, since when $\sum_{c=1}^C\sum_{k=1}^K{\boldsymbol{\nu}_{l,\{c,k\}}^T}{\bf y}_{l,\{c,k\}}=0$, at layer $l$, the c-sNTs \eqref{eq:nonlinear:representation} do not carry additional "information" different then the one in sNT \eqref{eq:sparse:representation}, while when $\boldsymbol{\nu}_{l,\{c,k\}}={\bf 0}$, the corresponding c-sNT reduces to the sNT.

\textit{\underline{$-$ Forward Weights Update}} 
Let all the variables in problem \eqref{eq:global:problem:formulation:reduction.0-1} be given except ${\bf A}_{l-1}$ 
then \eqref{eq:global:problem:formulation:reduction.0-1} 
reduces to the following problem: 
\begin{equation}
\begin{aligned}
\hat{\bf A}_{l-1}=\arg \min_{ {\bf A}_{l-1} } \Vert {\bf A}_{l-1}{\bf S}_{l-1}-{\bf W}_{l} \Vert_F^2+\mathcal{R}_2(l), \label{problem:decoupled:update:A}
\end{aligned}
\end{equation}
where we assume that:
\begin{equation}
\begin{aligned}
\! \! \! \! \! \! {\bf S}_{l-1}{\bf S}_{l-1}^T\!=\!{\bf U}_{l-1}\left( {\bf U}_{l-1}+\lambda_{l,b}\frac{\partial \mathcal{G}({\bf G}_{l-1},{\bf U}_{l-1})}{\partial {\bf U}_{l-1}} \right)^T \!\!\!, \\
\! \! \! \! \! \! {\bf W}_{l}\!=\!{\bf Y}_l-\frac{\partial \mathcal{G}({\bf G}_{l},{\bf U}_{l})}{\partial {\bf U}_{l}}-\lambda_{l,f}\frac{\partial \mathcal{G}({\bf G}_{l+1},{\bf U}_{l+1})}{\partial {\bf U}_{l+1}}.
\label{problem:decoupled:update:A:derivation}
\end{aligned}
\end{equation}
We give the derivation of \eqref{problem:decoupled:update:A:derivation} in \textit{Appendix} C. To solve \eqref{problem:decoupled:update:A} for ${\bf A}_{l-1} \in \Re^{M_{l} \times M_{l-1}}, M_{l} \geq M_{l-1}$, we use the approximate closed form solution of \cite{Kostadinov2018:EUVIP}. 
\textit{\underline{$-$ Backward Weights Update}}
Let all the variables in problem \eqref{eq:global:problem:formulation:reduction.0-1} be given except ${\bf B}_{l}$ 
then \eqref{eq:global:problem:formulation:reduction.0-1} reduces to the following problem:
\begin{equation}
\begin{aligned}
\! \! \! \! \! \! \! \! \hat{\bf B}_{l}\!=\!\arg \min_{ {\bf B}_{l} } \Vert {\bf B}_{l}{\bf U}_{l+1}\!-\!{\bf Y}_{l} \Vert_F^2 \! +\!\lambda_{l,5}\Vert {\bf B}_{l}-{\bf A}_{l}^T \Vert_F^2+\\
\lambda_{l,f}Tr\{ {\bf B}_{l}\frac{\partial \mathcal{G}({\bf G}_{l+1},{\bf U}_{l+1})}{\partial {\bf U}_{l+1}}\frac{\partial \mathcal{G}({\bf Q}_{l},{\bf Y}_{l})}{\partial {\bf Y}_{l}}^T \}, \label{problem:decoupled:update:B}
\end{aligned}
\end{equation}
which has a closed form solution as: 
\begin{equation}
\begin{aligned}
\! \! \! \! \! \! \hat{\bf B}_{l}\!=\!\left[ {\bf Y}_l{\bf U}_{l+1}^T+\lambda_{l,5}{\bf A}_{l}^T- \lambda_{l,f}\frac{\partial \mathcal{G}({\bf G}_{l+1},{\bf U}_{l+1})}{\partial {\bf U}_{l+1}}\right. \\
\! \! \! \! \! \! \left. \frac{\partial \mathcal{G}({\bf Q}_{l},{\bf Y}_{l})}{\partial {\bf Y}_{l}}^T\right]\left( {\bf U}_{l-1}{\bf U}_{l-1}^T+\lambda_{l,5}{\bf I} \right)^{-1}. \label{problem:decoupled:update:B.solution}
\end{aligned}
\end{equation}
The proof is straightforward by taking the first order derivative of \eqref{problem:decoupled:update:B} w.r.t. ${\bf B}_{l}$, equaling it to zero and reordering. If ${\bf B}_l={\bf A}_l^T$, then this step is omitted, while in \eqref{problem:decoupled:update:A:derivation} ${\bf B}_l$ is replaced by ${\bf A}_l^T$ that is estimated one iteration previously. 

\textbf{Local Convergence Guarantee for the Decoupled Problem} Note that for any of the decoupled problems \eqref{eq:global:problem:formulation:reduction.0-1}, in the estimation of the c-sNT representations, we have a closed form solution. In the forward weight ${\bf A}_{l-1}$ update,  we have an approximate closed form solution 
and in the backward weight ${\bf B}_{l}$ update, we have a closed form solution. Therefore, at each of the alternating steps, we have a guaranteed decrease of the objective $\sum_{j=1}^4{\mathcal{R}}_{j}(l)+{\mathcal{A}}(l)$, which allows us to prove a local gonvergence gourantee in similar fasion to the pfoof that is given by  \cite{Kostadinov2018:EUVIP}. 


%
%

\subsection{Synchronous and Asynchronous Execution}

Our algorithm has two possible execution setups.
In the first setup, a hold is active till all weights ${\bf A}_{l}$ and ${\bf B}_{l}$ 
in the network are updated by \textbf{Stage Two}. Afterwards, the execution of \textbf{Stage One} proceeds, which corresponds to a \textit{synchronous} case. 

In the second setup, at one point in time, one takes all the available weights ${\bf A}_{l}$ 
and ${\bf B}_{l}$, 
whether are updated or not in \textbf{Stage Two}, and executes \textbf{Stage One}, which corresponds to an \textit{asynchronous} case. In this way, the algorithm has the possibility to find a  solution to \eqref{eq:global:problem:formulation} by alternating between or executing in parallel \textbf{Stage One} and \textbf{Stage Two} under properly chosen scheduling scheme.

\subsection{Local Minimum Solution Guarantee}
The next result shows that with arbitrarily small error 
we can find a 
local minimum solution to 
\eqref{eq:global:problem:formulation} for ${\bf B}_l={\bf A}_l^T$. 
\textbf{Theorem 1}
\textit{Given any data set ${\bf Y}_0$, there exists} $\boldsymbol{\omega} =\{ \lambda_{1,bf}, ... , \lambda_{L,bf} \}, \lambda_{l, bf}=\{ \lambda_{l,b}, {\lambda_{l,f}}{} \}$
\textit{$\lambda_{l,b} >0, \lambda_{l,f}>0$ and a learning algorithm for a $L$-node transform-based network with a 
goal set on one 
node at level $l_G$ 
such that the 
algorithm 
after $t > S$ iteration 
learns all ${\bf A}_{l}, {l} \in \{0,...,L-1\}$ 
with}
\textit{
$\mathcal{G}({\bf D}_{L},{\bf U}_{L})  = {\epsilon}$, where  ${\bf D}_{L} \in \Re^{M_L \times CK}$ are the resulting representations 
of the propagated goal representations ${\bf G}_{l_G}$ through the network 
from node level $l_G+1$, and ${\epsilon}>0$ is arbitrarily small constant}. 

The proof is given in \textit{Appendix} D.

\textbf{Remark} 
\textbf{} \textit{ The result by \textbf{Theorem 1} reveals the possibility to attain 
desirable 
representations ${\bf U}_{L}$ at level $L$ 
while only setting one local representation goal on one node at level $l_G \in \{1,..,L \}$. } 

\vspace{-0.05in}

\section{Numerical Evaluation}



We present preliminary numerical evaluation of our learning strategy that is applied on a fully connected feed forward network, \textit{i.e.}, \eqref{eq:global:problem:formulation}, with square weights ${\bf A}_l, {\bf B}_l \in \Re^{N \times N}$, where ${\bf B}_l={\bf A}_l^T$. 

\vspace{-0.015in}
\subsection{Data, Evaluated NNs, and  Learning/Testing Setup} 
\vspace{-0.04in}
\textbf{Used Data and Evaluated Networks}  The used data sets are MNIST 
and Fashon-MNIST. 
All the images from the data sets are downscaled to resolution 
$28\times28$, and are normalized to unit variance. 
We analyze $12$ different networks, $6$ per database. Per one database $4$ networks have $6$ nodes and aditional $2$ have $4$ nodes. The networks are trained in synchronous \textit{syn} and asynchronous mode \textit{asyn}. For the $6$-node networks trained in \textit{syn}, $2$ of them have a goal defined at the last node $L$ (\textit{syn}$_{n[6]g[6]}$) and for the remaining $2$ the goal is set on node at the middle in the network at level $3$ (\textit{syn}$_{n[6]g[3]}$). For the $4$-node network the goal is set at node level $4$ (\textit{syn}$_{n[4]g[4]}$). Similarly for the \textit{asyn} mode, we denote the networks as (\textit{asyn}$_{n[6]g[6]}$), (\textit{asyn}$_{n[6]g[3]}$) and (\textit{asyn}$_{n[4]g[4]}$).      

\textbf{Scheduling Regime Setup for Network Learning} 
The asynchronous mode is implemented by using $L$ random draws $\boldsymbol{\phi} \in \{-1,1\}^{L}$, as the number of nodes, from a Bernoulli distribution. 
If the 
realization is $1$, $\phi(l)=1$, we use ${\bf A}_l^{t}$ in the forward pass (stage one) and we update the corresponding set of variables $\varsigma_l$ (stage two). If the 
realization is $-1$, $\phi(l)=-1$, then we do not use ${\bf A}_l^{t}$,
but, instead we use ${\bf A}_l^{t-1}$ for stage one and in stage two we do not update the corresponding set $\varsigma_l$. The synchronous mode is implemented by taking into account all ${\bf A}_l^t$.

An on-line variant is used for the update of ${\bf A}_{l}$ w.r.t. a subset of the available training set. It has the following form ${\bf A}^{t+1}_l={\bf A}^{t}_l-\rho({\bf A}^{t}_l-\hat{{\bf A}}_l )$, where $\hat{{\bf A}}_l$ and ${\bf A}^t_l$ are the solutions in the weight update step at iterations $t+1$ and  $t$, which is equivalent to having the additional constraint ${\Vert {\bf A}^t_l -{\bf A}_l^{t+1}\Vert_F^2}$ in the related problem and $\rho$ is a predefined step size 
(\textit{Appendix} C.1). 
The batch portion equals to $15\%$ of the total amount of the  training data. 
The parameters $\{\lambda_{l,1}, \lambda_{l,2}, \lambda_{l,3}, \lambda_{l,4}, \lambda_{l,5} \} = \{34, 34, 34, 34, 34 \}$ and $\lambda_{l,1}=M_l/(2 \times l)$.
All the parameters $\lambda_{l,fb}$ are set as $\lambda_{l,fb}=\{1, 1\}$. 
The algorithm is initialized with a random matrices having i.i.d. Gaussian (zero mean, unit variance) entries and is terminated after $120$ iterations. 

\vspace{-.04in}
\textbf{Evaluation Setup} 
All data are propagated through the learned network using the sNT \eqref{eq:sparse:representation}. 
Afterwords, the recognition results are obtained 
by using linear SVM \cite{Cortes:Vapnik:1995} on the network output representations. We take the corresponding training output network representations for learning the SVM and the testing output network representations for evaluation of the recognition accuracy. 

\begin{table}[t]
\center
\begin{minipage}[b]{0.66\linewidth}
\centering
\begin{tabular}{@{\hspace{-40pt}}c@{\hspace{1pt}}l|@{\hspace{1pt}}c@{\hspace{1pt}}c@{\hspace{1pt}}} 
\multicolumn{1}{c}{\bf } &  &\multicolumn{1}{c}{MNIST} &\multicolumn{1}{c}{F-MNIST} \\ 
\hline 
\multirow{5}{*}{Acc. [$\%$] }
&\text{state-of-the-art }   &\text{ $ $}\text{ $ $}\text{ $ $}\text{ $ $}\text{ $ $}\text{ $ $}$99.77^a$ &\text{ $ $}\text{ $ $}\text{ $ $}\text{ $ $}\text{ $ $}\text{ $ $}$94.65^b$ \\ 
& \textit{syn}$_{n[4]g[4]}$   &\text{ $ $}\text{ $ $}\text{ $ $}\text{ $ $}\text{ $ $}\text{ $ $}$98.75$ &\text{ $ $}\text{ $ $}\text{ $ $}\text{ $ $}\text{ $ $}\text{ $ $}$92.61$  \\ 
&\textit{syn}$_{n[6]g[6]}$   &\text{ $ $}\text{ $ $}\text{ $ $}\text{ $ $}\text{ $ $}\text{ $ $}$99.28$ &\text{ $ $}\text{ $ $}\text{ $ $}\text{ $ $}\text{ $ $}\text{ $ $}$93.23$ \\ 
&\textit{syn}$_{n[6]g[3]}$   &\text{ $ $}\text{ $ $}\text{ $ $}\text{ $ $}\text{ $ $}\text{ $ $}$98.98$ &\text{ $ $}\text{ $ $}\text{ $ $}\text{ $ $}\text{ $ $}\text{ $ $}$91.98$\\  
\\ 
\hline 
\multirow{4}{*}{Acc. [$\%$] } 
&\text{state-of-the-art}    &\text{ $ $}\text{ $ $}\text{ $ $}\text{ $ $}\text{ $ $}\text{ $ $}$99.77^a$ &\text{ $ $}\text{ $ $}\text{ $ $}\text{ $ $}\text{ $ $}\text{ $ $}$94.65^b$ \\ 
&\textit{asyn}$_{n[4]g[4]}$   &\text{ $ $}\text{ $ $}\text{ $ $}\text{ $ $}\text{ $ $}\text{ $ $}$98.57$ &\text{ $ $}\text{ $ $}\text{ $ $}\text{ $ $}\text{ $ $}\text{ $ $}$91.07$ \\ 
&\textit{asyn}$_{n[6]g[6]}$   &\text{ $ $}\text{ $ $}\text{ $ $}\text{ $ $}\text{ $ $}\text{ $ $}$99.01$ &\text{ $ $}\text{ $ $}\text{ $ $}\text{ $ $}\text{ $ $}\text{ $ $}$93.15$ \\ 
&\textit{asyn}$_{n[6]g[3]}$   &\text{ $ $}\text{ $ $}\text{ $ $}\text{ $ $}\text{ $ $}\text{ $ $}$98.91$ &\text{ $ $}\text{ $ $}\text{ $ $}\text{ $ $}\text{ $ $}\text{ $ $}$92.03$ \\ 
\end{tabular}
\end{minipage}
\caption{Comparative result for the recognition accuracy between the the feed-forward network learned using the proposed algorithm under synchronous and asynchronous update scheme and $^a$\cite{Schmidhuber:2012:MDN} and $^b$\cite{SaiSamarth:2018:DDCN}. }
\label{sample-table.sync-async}
\vspace{-.1in}
\end{table}
\begin{table}[t]
\vspace{-.21in}
\center
\begin{minipage}[b]{0.72\linewidth}
\centering
\begin{tabular}{@{\hspace{-40pt}}c@{\hspace{1pt}}l|@{\hspace{1pt}}c@{\hspace{1pt}}c@{\hspace{1pt}}} 
\\
\multicolumn{1}{c}{\bf } &  &\multicolumn{1}{c}{MNIST} &\multicolumn{1}{c}{F-MNIST}\\ 
\hline 
\multirow{3}{*}{t[h]} 
&\text{state-of-the-art}    &\text{ $ $}\text{ $ $}\text{ $ $}\text{ $ $}\text{ $ $}\text{ $ $}$14^a$ &\text{ $ $}\text{ $ $}\text{ $ $}\text{ $ $}\text{ $ $}\text{ $ $}$3^b$ \\ 
&\textit{syn}$_{n[6]g[6]}$   &\text{ $ $}\text{ $ $}\text{ $ $}\text{ $ $}\text{ $ $}\text{ $ $}$6 \times .5$ &\text{ $ $}\text{ $ $}\text{ $ $}\text{ $ $}\text{ $ $}\text{ $ $}$6 \times .5$ \\ 
&\textit{asyn}$_{n[6]g[6]}$   &\text{ $ $}\text{ $ $}\text{ $ $}\text{ $ $}\text{ $ $}\text{ $ $}$6 \times .6$ &\text{ $ $}\text{ $ $}\text{ $ $}\text{ $ $}\text{ $ $}\text{ $ $}$6 \times .7$ \\ 
\end{tabular}
\end{minipage}
\caption{Comparative result for the learning time in hours between the proposed algorithm under synchronous and asynchronous update scheme and $^a$\cite{Schmidhuber:2012:MDN} and $^b$\cite{SaiSamarth:2018:DDCN}.
}
\label{sample-table.sync-async.time}
\end{table}
\begin{table}[t]
\vspace{-.15in}
\center
\begin{minipage}[b]{0.89\linewidth}
\centering
\begin{tabular}{@{\hspace{-40pt}}c@{\hspace{1pt}}l|@{\hspace{1pt}}c@{\hspace{1pt}}c@{\hspace{1pt}}} 
\\
\multicolumn{1}{c}{\bf } &  &\multicolumn{1}{c}{MNIST} &\multicolumn{1}{c}{F-MNIST}\\ 
\hline 
\multirow{2}{*}{ \text{ }} 
&\text{Num. of connections}   &\text{ $ $}\text{ $ $}\text{ $ $}\text{ $ $}\text{ $ $}\text{ $ $}$6 \times N^2$ &\text{ $ $}\text{ $ $}\text{ $ $}\text{ $ $}\text{ $ $}\text{ $ $}$6 \times N^2$ \\ 
\end{tabular}
\end{minipage}
\caption{The size of the largest networks \textit{syn}$_{n[6]g[6]}$ and \textit{asyn}$_{n[6]g[6]}$ that were evaluated using our learning algorithm.}
\label{sample-table.sync-async.size}
\vspace{-.15in}
\end{table}

\subsection{Evaluation Summary} 
The results are shown in Tables \ref{sample-table.sync-async}, 
\ref{sample-table.sync-async.time} and \ref{sample-table.sync-async.size}. The networks trained using the proposed approach on both of the used databases achieve competative to state-of-the-art recognition performance w.r.t. results reported by  \citep{Schmidhuber:2012:MDN} and \citep{SaiSamarth:2018:DDCN}\footnote{For more details about the comparing network arhithecture as well as their learning time, we reffer to the original manuscripts \cite{Schmidhuber:2012:MDN} and \cite{SaiSamarth:2018:DDCN}}. 
More importantly, we point out that our networks have small number of parameters, \textit{i.e.}, $6$ networks with $6$ nodes having $6$ weights with dimensionality  $784 \times 784$ and $4$ networks with $4$ nodes having $4$ weights with dimensionality $784 \times 784$.  Whereas the learning time for $L=6$ node network is $ \sim 3.5$ hours, on a PC that has Intel® Xeon(R) 3.60GHz CPU and 32G RAM memory when using not optimized Matlab code that implements the sequential variant of the proposed algorithm. We expect a parallel implementation of the proposed algorithm to  provide $\sim L \times$ speedup, which would reduce the learning time to less then half an hour in our not optimized Matlab code.

\section{Conclusion}
In this paper, we introduced a novel learning problem formulation for estimating the network parameters.  
We presented insights, as well as unfolded new interpretations 
of the learning dynamics w.r.t. the proposed local propagation. 
We proposed a two stage learning strategy, which allows the network parameters to be updated in synchronous or asynchronous scheduling mode. We implemented it by an efficient algorithm that enables parallel execution of the learning stages. While in the first stage, our estimates are computed approximately, in the second stage, our estimates are computed exactly. Moreover, in the second stage, the solutions to the decoupled problems, have a local convergence guarantee. 

We showed theoretically that by learning with a local propagation constraint, we can achieve desired data propagation through the network  that enables attaining a targeted representations at the last node in the network. We empirically validated our approach. The preliminary numerical evaluation of the proposed learning principle was promising. On the used publicly available databases the feed-forward network trained using our learning principle provided comparable results w.r.t. the state-of-the-art methods, while having a small number of parameters and low computational cost.

The information-theoretic analysis on the fundamental limit in the trade-off between the local propagation, the local goal and the global data propagation flow as well as the study on "technical" goals, e.g., goals that add to the acceleration in convergence of the learning is one future direction. Performance evaluation on other and large data sets, together with comparative evaluation for other activation functions, goals (e.g., reconstruction, discrimination, robustness, compression, privacy and security related goals) or a combination of them under different penalties $\psi$, is another future direction.

We point out that for other network architectures as well as for multi-path network a similar problem formulation could be considered. Moreover, by adopting the presented approach, similar solutions could also be derived. 
In fact, our algorithm, is applicable for network defined as a directed graph, \textit{i.e.}, a network where the propagation flow is specified and known. 

The proposed learning principle allows us by only changing the constraints on the propagation flow to influence on the properties of all hidden and output representations. In this line, the next frontier towards the ultimate machine intelligence could be seen in unsupervised self-driven goals, propagation flows and self-configuration. Where the network will learn what will be the goals, what kind of constraints on the propagation flow is required to reach that goal and how many nodes are required. 

\bibliography{icml2019}
\bibliographystyle{plainnat}
\end{document}